# Use Image Clustering to Facilitate Technology Assisted Review

Haozhen Zhao
Data & Technology
Ankura Consulting Group, LLC
Washington DC, USA
haozhen.zhao@ankura.com

Fusheng Wei
Data & Technology
Ankura Consulting Group, LLC
Washington DC, USA
fusheng.wei@ankura.com

Hilary Quatinetz
Data & Technology
Ankura Consulting Group, LLC
Washington DC, USA
hilary.quatinetz@ankura.com

Han Qin
Data & Technology
Ankura Consulting Group, LLC
Washington DC, USA
han.qin@ankura.com

Adam Dabrowski
Data & Technology
Ankura Consulting Group, LLC
Washington DC, USA
adam.dabrowski@ankura.com

*Abstract—* **During the past decade breakthroughs in GPU hardware and deep neural networks technologies have revolutionized the field of computer vision, making image analytical potentials accessible to a range of real-world applications. Technology Assisted Review (TAR) in electronic discovery though traditionally has dominantly dealt with textual content, is witnessing a rising need to incorporate multimedia content in the scope. We have developed innovative image analytics applications for TAR in the past years, such as image classification, image clustering, and object detection, etc. In this paper, we discuss the use of image clustering applications to facilitate TAR based on our experiences in serving clients. We describe our general workflow on leveraging image clustering in tasks and use statistics from real projects to showcase the effectiveness of using image clustering in TAR. We also summarize lessons learned and best practices on using image clustering in TAR.**

*Keywords- predictive coding, legal document review, technology assisted review, deep learning, image clustering*

## I. INTRODUCTION

In the legal industry, to reduce cost and time spent in electronic discovery, it has become a common practice to use machine learning for document review, known as predictive coding or TAR (technology assisted review). Both traditional machine learning methods and deep learning algorithms have been studied for tasks such as text classification and document clustering. [2, 3]. However, these applications only deal with text documents. Images, sounds, videos files in the document depository are usually left out for eyes-on review or unreviewed, which would have cost implications if these types of documents made up more than a small fraction of the total review set [1].

With the advance of computer vision with deep learning and transfer learning, applying machine learning to TAR for visual documents has become a reality. Authors in [5] presented several business cases arise from real legal document review projects for applications of image analytics, which include applications of image classification, image clustering, and object detection. These applications use transfer learning approach to take advantage of well-established pretrained deep learning models.

One of the applications, the Image Clustering, has seen increasing usages by various projects and gain popularity with users. In this paper, we present in-depth several business cases to showcase the effectiveness and benefits of using image clustering in TAR for image documents.

This paper is organized as following. Section II covers the underlying technologies used in the image clustering application and its overall architecture. Section III presents two recent business cases powered by the image clustering application. Section IV concludes the paper with lessons we learned from using image clustering in TAR and remarks on potential related future work.

## II. IMAGE CLUSTERING

The image clustering application supporting the business cases was initially presented in [5]. At its core, the application uses K-means clustering algorithms on dense features extracted with pretrained model VGG16 [4]. Specifically, it extracts the feature vectors from the second fully connected layer ("fc2", the last layer before the classification layer) of the VGG16 model. The feature vectors have a dimension of 4095, which are used as input for the K-means clustering algorithm. The application is deployed to the production environment with a front-end Web GUI, which provides functionalities for users to select and tag images for clustering, to explore and interact with the clustering results. The application is integrated with the Relativity discovery platform.

Since the publication of paper [5], we have made additional performance enhancements and added new features to the application to meet the demands of growing number of projects with large set of images.

On the performance enhancements, we use parallel processing on multiple GPUs to speed up the feature extraction with VGG16. We also added preprocessing steps to deduplicate images in the file system to avoid repeated extraction from the identical images. Since the VGG16 model is applied in batches, some bad formatted images can cause a whole batch to fail. We addressed the issue with image preprocessing and recursive application of the model to extract features for all valid images.

A new feature added to the image clustering application is to find similar images of a given image, independent of the clusters. We originally used similarity matrix approach, that is, to compute pairwise Euclidean distances with the extracted feature vectors, then for each image, to find the top k images that are closest to the given image. Since the feature vectors have a dimension of 4096, it soon ran into two problems – one is that the execution time is too long, often in hours; and the other it runs out of memory if the image set is too large.

To address these issues, we turned to the Approximate Nearest Neighborhood (ANN) approach. We use the library Flann [6], which is very fast and has a good precision in the nearest neighbor search; and it is scalable. In a comparison test with the similarity matrix approach for 50 nearest neighbors, we found the precision is high for about the first half closest neighbors, whereas the further neighbors are not as precise as those of the similarity matrix.

Fast nearest neighborhood search with large size of high dimensional data is an increasingly important problem. Application areas include computer vision and recommender systems. There are many ANN models and there are benchmarks for various ANN models [7]. We plan to evaluate some more models to find the best ones to enhance our application.

### III. BUSINESS CASES USING IMAGE CLUSTERING

Since the deployment of the image clustering application, we have used it in a range of client projects. Below we select two recent representative projects showcasing how the application was used in practice and the value it created for clients in the cases.

#### A. Project A

For Project A. counsel had 4 weeks to collect, process, review, and produce to the FTC for a 2$^{nd}$ request matter. We leveraged Ankura's Predict technology to reduce the volume of documents needing manual review. Further, our client also needed to consider how to review image files that were excluded from TAR as they expected a subset of the images would be responsive to the FTC specification requests. There were a total of ~190k image files within the review universe. Ankura was able to leverage the Image Cluster analytics tool to significantly reduce the amount of time needed to review the 190K image files.

For Project A, we ran two rounds of image clustering and created 100 distinct clusters for each round. For each round, we estimate 5 hours for Ankura to identify, prepare, review and categorize the image clusters. The review time for counsel to designate clusters as Responsive, Not Responsive, or Further Review was 2 hours for each round. By reviewing the visual clusters where the image files are displayed in thumbnail view, counsel could view the clusters with likely substantive images very quickly, with hundreds and sometimes thousands of images in a matter of minutes. Counsel relied on the cluster categorization report Ankura created to determine which clusters needed review. Clusters of headshots, logos, icons, emojis, photos of animals, phone contacts, wordart, memes are some examples of generic clusters that were quickly determined to be non-responsive. 131,564 documents were determined to be non responsive and 56,793 documents were determined to be responsive with little attorney review time. Only 808 image files required manual review after this cluster analysis workflow. Our client was thrilled with how much time using image clustering saved them, which we estimate at 400 to 500 hours of contract review time.

If time permitted, we could have condensed the clustering into one round, but the client faced an incredibly accelerated production timeline and needed to run the first round before all data was added to the review universe in order to start rolling productions and meet the FTC deadline. The use of image clustering for Project A was a critical factor in ensuring the client met the tight production deadline.

#### B. Project B

For Project B, counsel had to meet an accelerated timeline to produce to government regulatory agencies. Similar to Project A, this was a 2$^{nd}$ request matter and they leveraged a TAR review strategy here as well. Counsel leveraged image clustering to identify images containing potential privileged content or Personal Identifiable Information (PII). We ran one round of image clustering across 38,957 image files and created 150 distinct clusters. We estimate 5 hours for Ankura to identify, prepare, review and categorize the image clusters. The categorization report included labels for images containing PII, such as social security cards. Four clusters were identified as containing PII. Counsel did not review any of the 150 image clusters and instead made a determination to produce all images except those with clusters containing PII. We estimate that the client saved about 75 to 100 hours of contract review time if they reviewed these images manually.

## IV. LESSONS AND BEST PRACTICES

We discuss below the lessons we learned in using image clustering. In particular, we find that deciding an optimal number of clusters in clustering is often challenging and preliminary analysis to exclude easy non responsive image before running the clustering is always helpful.

After running numerous iterations of image clustering across various data sets of different sizes and sources, we have determined that setting the number of clusters at 150 generates the best results. It would be reasonable to assume that a smaller number of images might indicate less clusters are necessary. However, we have learned that in order to get the most unique categories of images grouped together it is recommended to set the number of clusters at 150 regardless of image count. If there is not enough clusters generated, there is a risk of creating clusters that contain images with different enough characteristics that reviewers could potentially spend more time reviewing these clusters with mixed results than the time it would take to review additional unique clusters. Ultimately, clients are better off reviewing more clusters that contain unique characteristics as it will be easier to categorize the results and make a decision on whether the content of those clusters are responsive, not responsive, or require review. Ensuring the clusters contain this richness in uniqueness will also result in more accurate responsive or non-responsive coding decisions.

Performing a preliminary analysis on the image set to identify and exclude high frequency non responsive images from the image clustering set is another best practice we recommend. This can be done through tallying and sorting MD5HASH, filesize, or other image file properties to identify the highest frequency images and review them to confirm content is not responsive. Usually, the highest frequency images are logos that can be easily coded as not responsive and can be excluded from the image clustering analysis to be run.